\documentclass[a4paper]{esannV2}
\usepackage[dvips]{graphicx}
\usepackage[latin1]{inputenc}
\usepackage{amssymb,amsmath,array}
\usepackage{hyperref}
\usepackage{longtable}
\usepackage{makecell}
\usepackage{pdfpages}
\usepackage{subcaption}
\usepackage{amsthm}
\usepackage{sidecap}

\theoremstyle{definition}
\newtheorem{exmp}{Example}

\begin{document}
\title{Impact of Biases in Big Data}

\author{Patrick Glauner$^1$, Petko Valtchev$^{2}$ and Radu State$^1$
%
%
\vspace{.3cm}\\
%
1- Interdisciplinary Centre for Security, Reliability and Trust, \\ University of Luxembourg \\
29, Avenue JF Kennedy, 1855 Luxembourg, Luxembourg
%
\vspace{.1cm}\\
2- Department of Computer Science, University of Quebec in Montreal \\
201, av. President Kennedy, Montreal H2X 3Y7, Canada
}
\setlength{\tabcolsep}{0pt}

\maketitle

\begin{abstract}
The underlying paradigm of big data-driven machine learning reflects the desire of deriving better conclusions from simply analyzing more data, without the necessity of looking at theory and models. Is having simply more data always helpful? In 1936, The Literary Digest collected 2.3M filled in questionnaires to predict the outcome of that year's US presidential election. The outcome of this big data prediction proved to be entirely wrong, whereas George Gallup only needed 3K handpicked people to make an accurate prediction. Generally, biases occur in machine learning whenever the distributions of training set and test set are different. In this work, we provide a review of different sorts of biases in (big) data sets in machine learning. We provide definitions and discussions of the most commonly appearing biases in machine learning: class imbalance and covariate shift. We also show how these biases can be quantified and corrected. This work is an introductory text for both researchers and practitioners to become more aware of this topic and thus to derive more reliable models for their learning problems.
\end{abstract}

\section{Introduction}
\label{section:intro}
For about the last decade, the Big Data paradigm that has dominated research in machine learning can be summarized as follows: ``It's not who has the best algorithm that wins. It's who has the most data." \cite{banko2001scaling} In practice, however, most data sets are (systematically) biased.

\begin{exmp}
A spam filter is trained on a data set that consists of positive and negative examples. However, that training set was created a few years ago. Recent spam emails are different in two ways: the content of spam emails is different and the proportion of spam among all emails sent out has changed. As an outcome, the spam filter does not detect spam reliably and becomes even less reliable over time.
\end{exmp}

The appearance of biases in data sets imply a number of severe consequences including, but not limited to, the following: First, conclusions derived from biased - and therefore unrepresentative - data sets could simply be wrong due to lack of reproducibility and lack of generalizability. This is a common issue in research as a whole, as it has been argued that most research published may actually be wrong \cite{ioannidis2005most}. Second, these machine learning models may discriminate against subjects of under-represented categories \cite{curtisgoogle2015, wang2017deep}.

\begin{figure}[h!]
    \centering
    \begin{subfigure}[b]{0.49\textwidth}
    \includegraphics[width=\textwidth]{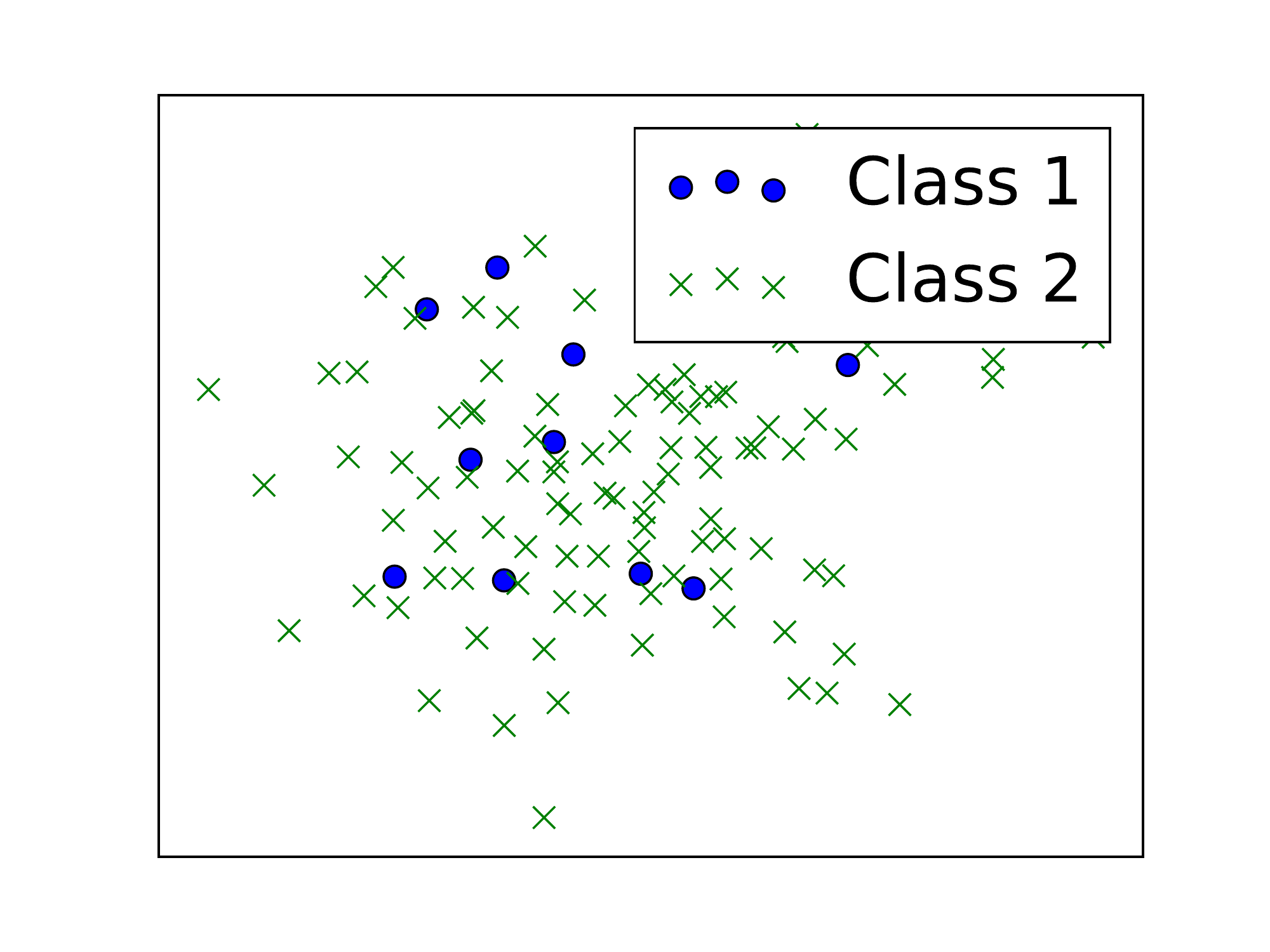}
    \caption{Class imbalance: Classes are unequally represented in the data.}
    \label{fig:imbalance}
    \end{subfigure}
    \begin{subfigure}[b]{0.49\textwidth}
    \includegraphics[width=\textwidth]{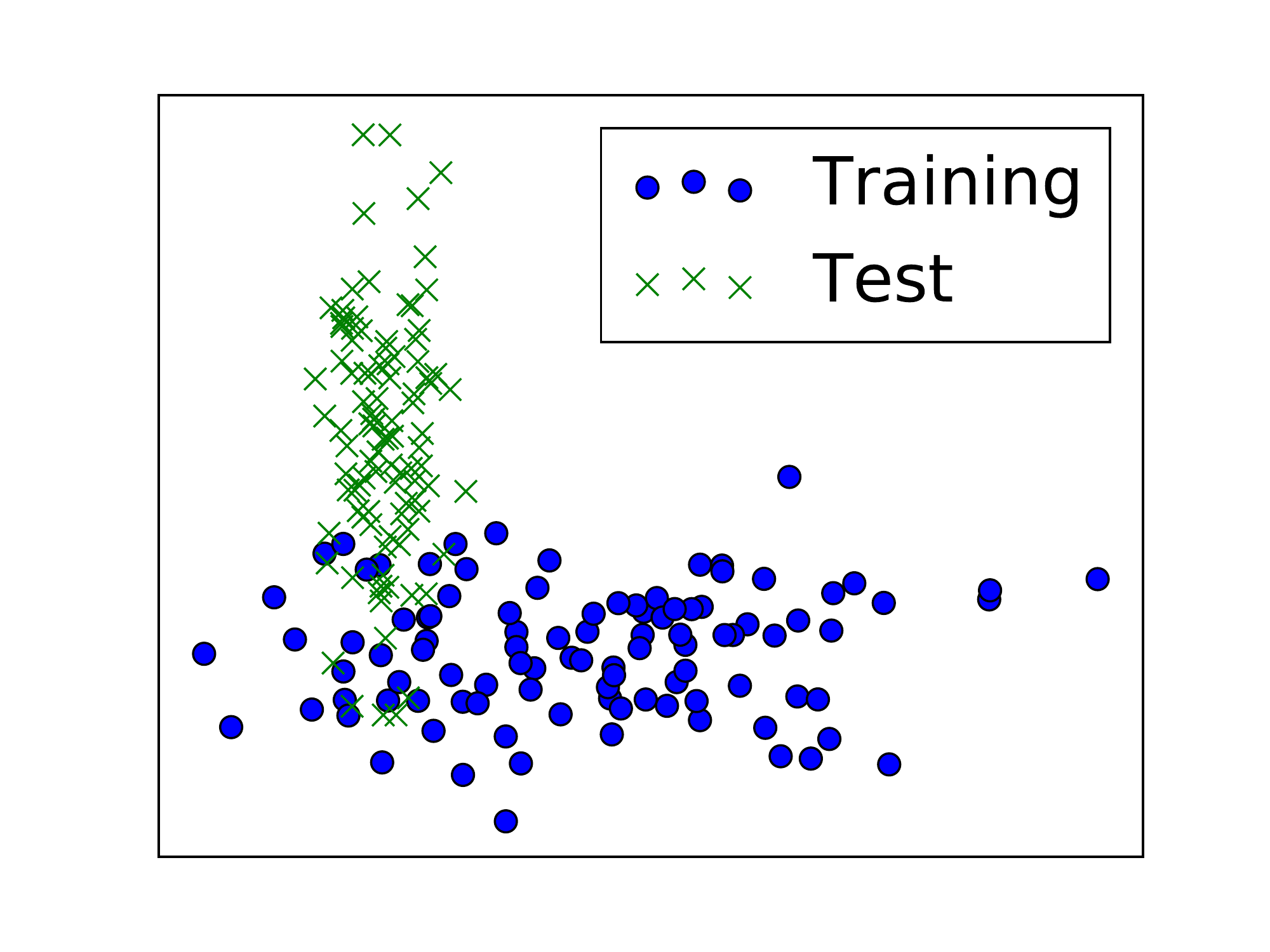}
    \caption{Covariate shift: Training and test data sets are drawn from different distributions.}
    \label{fig:cov}
    \end{subfigure}
    \caption{The most commonly appearing biases in data sets.}
    \label{fig:biases}
\end{figure}

From a technical perspective, the most commonly appearing biases include \textit{class imbalance} and \textit{covariate shift}. Class imbalance is the case where classes are unequally represented in the data. An example is visualized in Fig.~\ref{fig:imbalance}. Covariate shift is the problem of drawing training and test data sets from different distributions. An example is visualized in Fig.~\ref{fig:cov}. 
These biases are often ignored in both research and practical applications.
In part of the statistical literature, the phenomenon of biased data sets is called non-stationarity. In essence, this term indicates different statistics at a different time of collection of the training and test data sets, respectively \cite{sayed2012learning}.

More generally, however, the term \textit{bias} is multifaceted in the field of machine learning and describes different matters: The inductive bias of a learning algorithm refers to the set of assumptions a learner makes \cite{mitchell1997machine}. For example, logistic regression assumes that the training data is linearly separable. In contrast, the term bias is often used as a synonym for underfitting in the literature \cite{bishop2006pattern}.
Moreover, the parameter $w_0$ of a hypothesis
\begin{align*}
h(x) = w_0 + w_1 x_1 + ... + w_n x_n
\end{align*}
is sometimes called bias as it allows to shift a hypothesis by a fixed offset \cite{bishop2006pattern}. 

Our core contribution in this work is that we provide a systematic review of the research on biased data sets in the field of machine learning. We show that Big Data is useful, but not a silver bullet that could simply always be used without correcting the biases in it.
The results of this review allow researchers to pay attention to this topic in their own research and thus leading to more reliable models. The rest of this paper is organized as follows: Section~\ref{section:motivation} provides a selection of popularized issues caused by applications of machine learning models trained on biased data sets. Section~\ref{section:bias} provides a general introduction of how biases in data sets are defined. Sections~\ref{section:imbalance} and ~\ref{section:cov} provide reviews of class imbalance and covariate shift - the most commonly appearing biases in data sets, respectively. Section~\ref{section:other} provides references to other biases in data sets studied in the literature. Section~\ref{section:conclusions} summarizes this work.

\section{The more data, the better?}
\label{section:motivation}
Historically, biased data sets have been a long-standing issue in statistics. The following example describes the failed prediction of the outcome of the 1936 US presidential election. It is often cited in the statistics literature in order to illustrate the impact of biases in data. This example is discussed in detail in \cite{bryson1976literary}.

\begin{exmp}
The Democratic candidate Franklin D. Roosevelt was elected President in 1932 and ran for a second term in 1936. Roosevelt's Republican opponent was Kansas Governor Alfred Landon. \textit{The Literary Digest}, a general interest weekly magazine, had correctly predicted the outcomes of the elections in 1916, 1920, 1924, 1928 and 1932 based on straw polls. In 1936, The Literary Digest sent out 10M questionnaires in order to predict the outcome of the presidential election. The Literary Digest received 2.3M returns and predicted Landon to win by a landslide. However, the predicted result proved to be wrong, as quite the opposite happened: Roosevelt won by a landslide. This leads to the following questions:
\begin{enumerate}
\item How could the prediction turn out to be completely wrong despite the 2.3M participants?
\item How could The Literary Digest actually collect 10M addresses in 1936?
\end{enumerate}
The Literary Digest compiled their data set of 10M recipients mainly from car registrations and phone directories. In that time, the households that had a car or a phone represented a disproportionally rich, and thus biased, sample of the overall population that particularly favored the Republican candidate Landon. In contrast, George Gallup only interviewed 3K handpicked people, which were an unbiased sample of the population. As a consequence, Gallup could predict the outcome of the election very accurately \cite{harford2014big}.
\end{exmp}

Even though this historic example is well understood in statistics nowadays, similar or related issues happen every day dozens of times in modern Big Data-oriented machine learning. We now discuss selected examples that result from biases in modern applications of Big Data-driven machine learning.

\begin{exmp}
It has been argued that most data on humans may be on white people and thus may not represent the overall population \cite{united2014big}. As a consequence, the predictions of models trained on such biased data may cause infamous news. For example, in 2015, Google added an auto-tagging feature to its Photos app. This new feature automatically assigned tags to photos, such as bicycle, dog, etc. However, some black users reported that they were tagged as ``gorillas", which led to major criticism of Google \cite{curtisgoogle2015}. Most likely, this mishap was caused by a biased training set, in which black people were largely underrepresented.
\end{exmp}

The examples provided in this Section show that having simply more data is not always helpful in training reliable models, as the data sets used may be biased. In the following Sections, we discuss the most commonly appearing biases in data sets. We also present different strategies for assessing biased models and how to correct biases. These techniques include weighting training examples as well as subsampling methods. As a consequence, having data that is more representative is favorable, even if the amount of data used is less than just using the examples from a strongly biased data set.

\section{Biases in data sets}
\label{section:bias}
In supervised learning, training examples $(x^{(i)}, y^{(i)})$ are drawn from a training distribution $P_{train}(X, Y)$, where $X$ denotes the data and $Y$ the label, respectively. The training set is biased if the following inequality holds true:
\begin{align}
P_{train}(X, Y)\ne P_{test}(X, Y).
\end{align}
Different biases are visualized in Fig.~\ref{fig:biases}. In order to reduce a bias, it has been shown that example $(x^{(i)}, y^{(i)})$ can be weighted during training as follows \cite{japkowicz2002class}:
\begin{align}
\label{eq:weight}
w_i = \frac{P_{test}(x^{(i)}, y^{(i)})}{P_{train}(x^{(i)}, y^{(i)})}.
\end{align}

However, computing $P_{train}(x^{(i)}, y^{(i)})$ may be impractical in many cases because of the limited amount of data in the training domain. In the following sections, we discuss different biases for which specific assumptions about $P_{train}(X, Y)$ and $P_{test}(X, Y)$ are made.

\section{Class imbalance}
\label{section:imbalance}
Class imbalance refers to the case where classes are unequally represented in the data. When comparing training set and test set, respectively, we assume \cite{jiang2008literature}:
\begin{align}
P_{train}(Y)&\ne P_{test}(Y), \\
P_{train}(X\lvert Y) &= P_{test}(X\lvert Y).
\end{align}

An example is depicted in Fig.~\ref{fig:imbalance}. Imbalanced classes appear frequently in machine learning. Machine learning models trained on imbalanced data sets often tend to predict the majority class. The appearance of imbalanced classes also affects the choice of evaluation metric. Accuracy and recall are the most commonly used metric in contemporary research works in machine learning \cite{japkowicz2002class, tang2009svms}.
However, both metrics are affected by class imbalance. As a consequence, in many machine learning works, overly high accuracies or recalls are reported \cite{glauner2017challenge}.

\begin{exmp}
Anomaly detection problems often work on particularly imbalanced data sets. A test set containing 1K customers of which 999 have regular behavior and 1 has irregular behavior, (1) a classifier always predicting regular behavior has an accuracy of 99.9\%, whereas in contrast, (2) a classifier always predicting irregular behavior has a recall of 100\%. While the classifier of the first example has a very high accuracy and intuitively seems to perform very well, it will never predict any irregular behavior. In contrast, the classifier of the second example will find all customers that have irregular behavior, but may potentially trigger many costly and unnecessary interventions for customers that have a regular behavior \cite{glauner2017challenge}.
\end{exmp}

\begin{exmp}
The Modified National Institute of Standards and Technology (MNIST) database consists of 60K training images and 10K test images used for recognition of hand-written digits \cite{lecun1998gradient}, for which examples are depicted in Fig.~\ref{fig:mnistexample}. MNIST has been used in the fields of computer vision and machine learning for the last 20 years. The test accuracies reported in recent research are above 99.6\% \cite{wan2013regularization, sabour2017dynamic}. The distribution of test labels is depicted in Fig.~\ref{fig:mnistimbalance:test}. We notice that this data set is mainly imbalanced between the different classes. As a consequence, the accuracy is not the right metric for MNIST, as an increase of this metric does not necessarily imply an increased predictive power of a model. We would like to add that the distribution of labels is nearly the same for the training set. Furthermore, there is another imbalance between the training set and test set, respectively. However, that imbalance is less noticeable and we have therefore focused on the imbalance between the labels in each set.

\begin{figure}[h!]
    \centering
    \begin{subfigure}[b]{0.39\textwidth}
    \centering
    \includegraphics[width=0.5\textwidth]{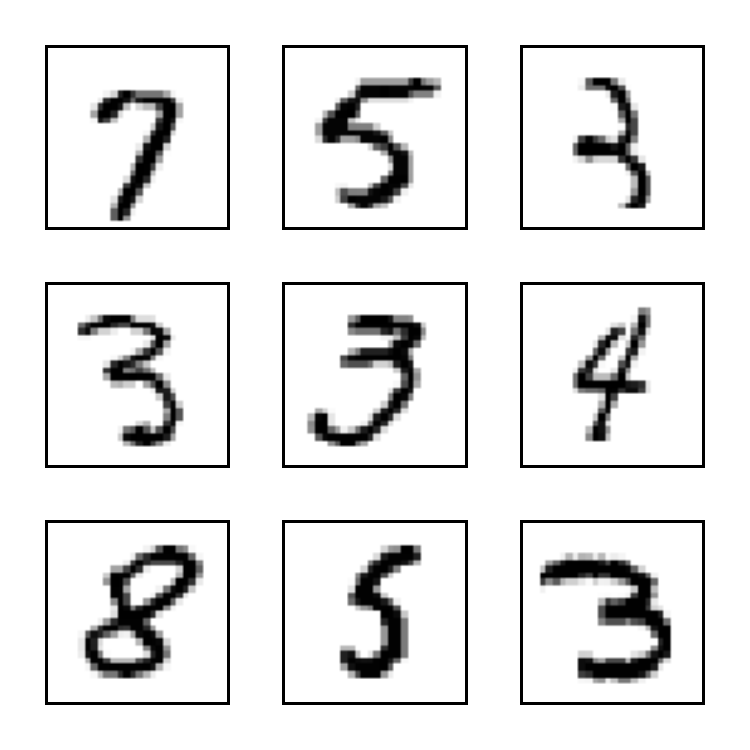}
    \caption{Example images.}
    \label{fig:mnistexample}
    \end{subfigure}
    \begin{subfigure}[b]{0.54\textwidth}
    \centering
    \includegraphics[width=0.5\textwidth]{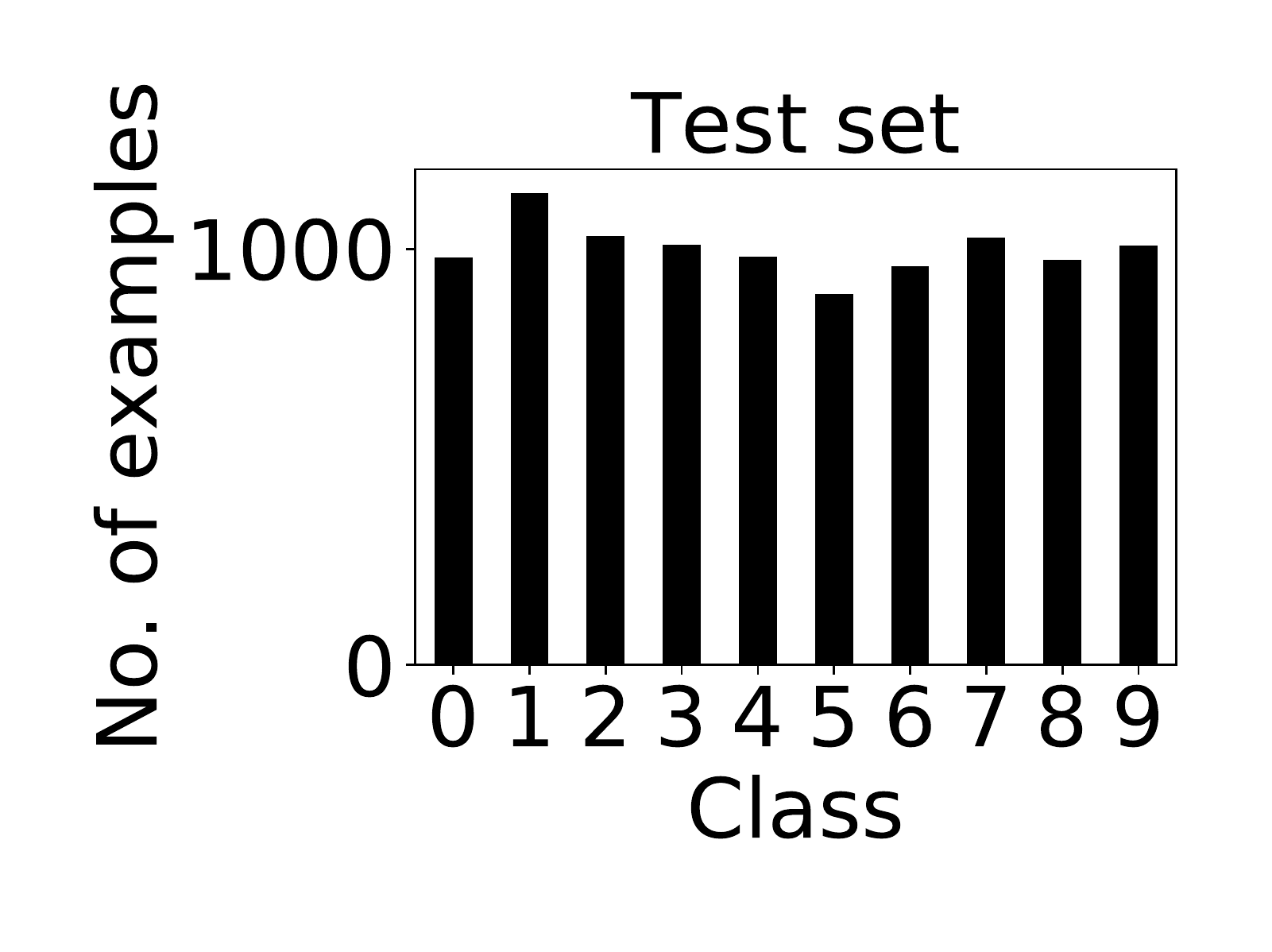}
    \caption{Distribution of test labels.}
    \label{fig:mnistimbalance:test}
    \end{subfigure}
    \caption{MNIST data set.}
    \label{fig:mnist}
\end{figure}
\end{exmp}

\begin{figure}
    \centering
    \includegraphics[width=0.4\textwidth]{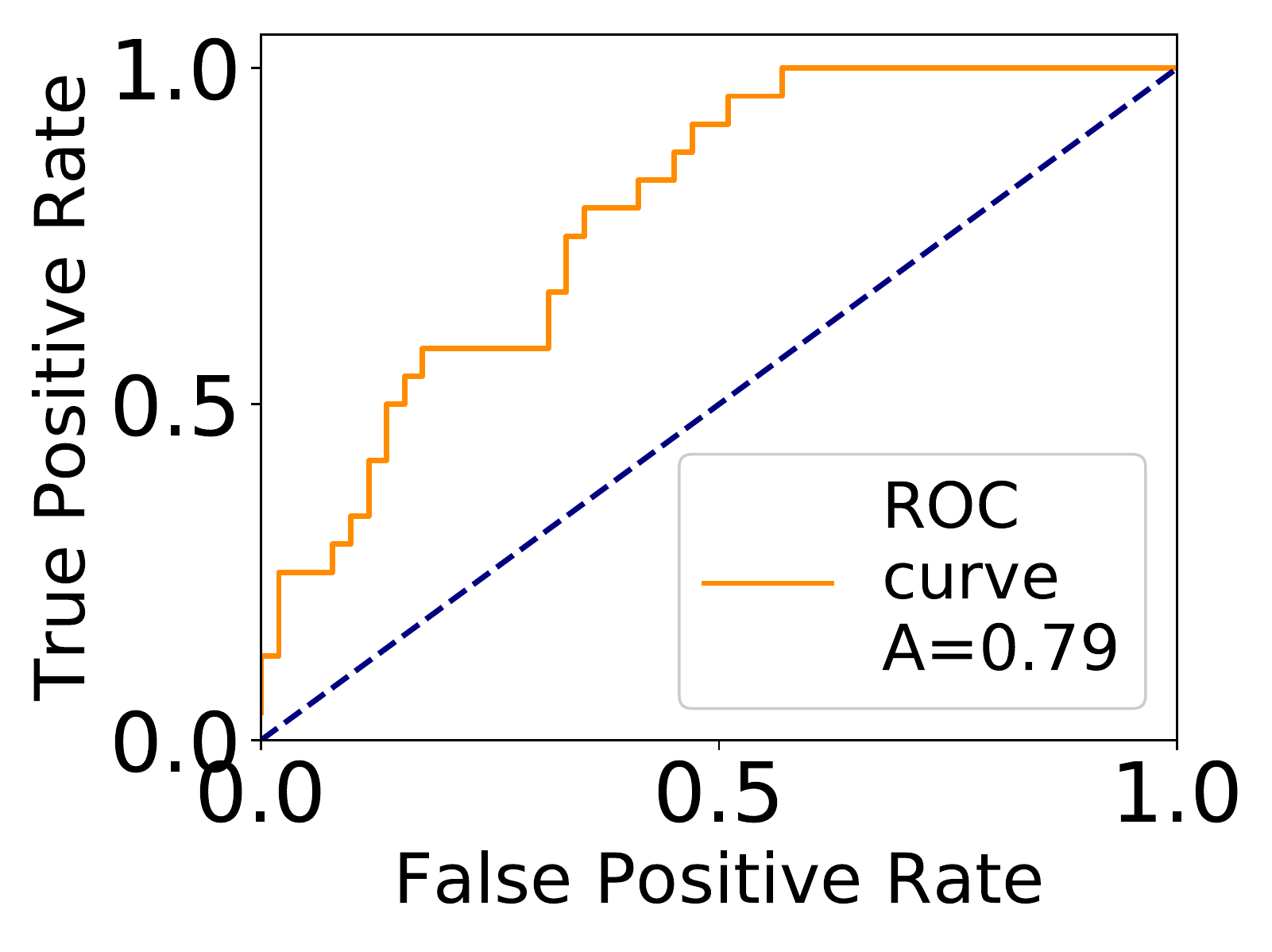}
    \caption{Example of receiver operating characteristic (ROC) curve.}
    \label{fig:rocexample}
\end{figure}

A number of metrics that are insensitive to class imbalance can be found in the literature. One common metric is to use a receiver operating characteristic (ROC) curve, which plots the true positive rate against the false positive rate for varying decision threshold values. An example is depicted in Figure~\ref{fig:rocexample}. The area under the curve (AUC) is a performance
measure between 0 and 1, where any binary classifier with an AUC $>0.5$ performs better than random guessing \cite{fawcett2006introduction}.
Another metric that is insensitive to class imbalance is the Matthews correlation coefficient (MCC):
\begin{align}
 \frac{TP\times TN - FP\times FN}{\sqrt{(TP + FP)(TP+FN)(TN+FP)(TN+FN)}}, \label{eq:MCC}
\end{align}
which measures the accuracy of binary classifiers taking into account the imbalance of both classes, ranging from $-1$ to $+1$ \cite{matthews1975comparison}.
Furthermore, for multi-class problems the intraclass correlation coefficient (ICC) has been proposed \cite{stanish1983estimation}. It can be interpreted as the fraction of the total variance that is between the different classes. It has been  successfully applied to imbalanced multi-class learning problems \cite{werner2015handling}.

In order to correct the class imbalance during training, a number of methods are proposed in the literature. First, weighting examples by the inverse proportion of examples per class using Eq.~\ref{eq:weight} is proposed in the literature \cite{jiang2008literature}.
On the one hand, one intuitive method is undersampling the majority classes by dropping training examples, either randomly or by specific criteria \cite{tomek1976two, mani2003knn}. This approach leads to smaller data sets, but may lack variation, as important examples could have been dropped. On the other hand, oversampling the minority classes by creating more training examples is proposed in the literature. Most trivially, training examples can simply be randomly copied. However, there are also more sophisticated algorithms, such as the synthetic minority over-sampling technique (SMOTE), which attempts to create synthetic examples representing the minority class by interpolating between neighboring data points \cite{chawla2002smote}. Generally, adding more examples leads to larger training sets, which, in turn, leads to increased training time. Therefore, combinations of oversampling and undersampling were proposed \cite{batista2003balancing, liu2009exploratory}.

\section{Covariate shift}
\label{section:cov}
Covariate shift refers to the case where the training data and test data are distributed differently. We assume \cite{jiang2008literature}:
\begin{align}
P_{train}(X)&\ne P_{test}(X), \label{eq:cov1} \\
P_{train}(Y\lvert X) &= P_{test}(Y\lvert X). \label{eq:cov2}
\end{align}

An example is depicted in Fig.~\ref{fig:cov}. Covariate shift appears frequently in machine learning as discussed in Section~\ref{section:motivation}. Machine learning models trained on biased training sets tend not to generalize on test data that is from the true underlying distribution of the population.

\begin{exmp}
Non-technical losses (NTL) appear in power grids during distribution and describe irregular power usage, in particular electricity theft. NTL are reported to range up to 40\% of the total electricity distributed in countries such as Brazil, India, Malaysia or Pakistan \cite{glauner2017challenge, viegas2017solutions}. Recent research on NTL detection mainly uses machine learning models that learn anomalous behavior from customer data and known irregular behavior that was reported through on-site inspection results. We have previously shown that in many cases, the set of inspected customers is biased as depicted in Fig.~\ref{fig:NTLexample} \cite{2017glaunerisbigdata}. A reason for this bias is that past inspections have been largely focused on certain criteria and were not sufficiently spread across the population. As a consequence, when learning from the inspection results, a bias is learned, making predictions less reliable.

\begin{figure}
\centering
\includegraphics[width=0.4\textwidth]{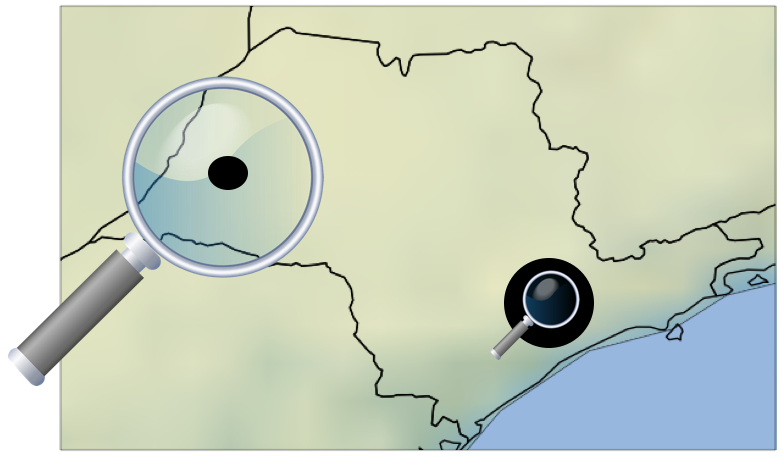}
\caption{Example of spatial bias: Most inspections are carried out in the small city due to hidden selection criteria. This sample of customers inspected thus does not represent the overall population of customers \cite{2017glaunerisbigdata}.}
\label{fig:NTLexample}
\end{figure}
\end{exmp}

The literature distinguishes classifiers into local learners and global learners, respectively \cite{zadrozny2004learning}. For a local learner, the prediction of the learner depends only on $P(Y\lvert X)$ for an increasing number of training examples. Previous research addresses this behavior with the term ``asymptotically". We assume $P_{train}(Y\lvert X) = P_{test}(Y\lvert X)$ in Eq.~\ref{eq:cov2}. Hence, a local learner is not affected by covariate shift. Examples include logistic regression and hard-margin support vector machine (SVM).
In contrast, the prediction of a global learner depends asymptotically on both, $P(Y\lvert X)$ and $P(X)$. We assume $P_{train}(X) \ne P_{test}(X)$ in Eq.~\ref{eq:cov1}. Hence, a global learner is affected by covariate shift.
Examples include decision tree learners such as ID3 or C4.5, naive Bayes and soft-margin SVM \cite{2017glaunerisbigdata}.
The terms ``global" and ``local'', respectively, have been established as follows: A global learner also uses $P(X)$, which is a (global) distribution over the entire input data. In contrast, a local learner uses $P(Y\lvert X)$, which refers for every $x^{(i)}\in X$ to a local distribution $P(Y\lvert x^{(i)})$.

When using a data set, we need to assess whether the training set has actually a covariate shift. The Kullback-Leibler divergence \cite{kullback1987letter} is a measure of the difference of two probability distributions. However, it is challenging (1) to adapt this measure to multi-dimensional data that is a combination of discrete and continuous features, which is common in machine learning, and (2) to define criteria from what values on a distance is an indicator for a covariate shift. We have recently proposed decision tree learning for finding a model that is able to distinguish between training and test distributions. The rationale behind our methodology is as follows: First, we add a feature $s$ and assign the values $1$ or $0$ to the training data $(s = 1)$ or test/production data $(s = 0)$, respectively. Second, decision trees are global learners and thus sensitive to covariate shift. Third, optimizing a decision tree to predict the label $s$ and thus maximizing this distinction between the two sets is equivalent to finding the best binary classification between test/production data and original training data. Next, the performance of the classifier is quantified using the Matthews correlation coefficient (MCC), which is defined in Eq.~\ref{eq:MCC}. As a result, the MCC value is the magnitude of the covariate shift in the data set \cite{2017glaunerisbigdata}.

Instance weighting using density estimation has been proposed for correcting covariate shift \cite{shimodaira2000improving}. Examples can either be weighted during training \cite{cortes2014domain} or the weights can be used for rejection sampling \cite{zadrozny2004learning}.
Historically, the Heckman method has been proposed to correct covariate shift by estimating the probability of an example being selected into the training sample \cite{10.2307/1912352}. However, the Heckman method only applies to linear regression models.

\section{Other biases}
\label{section:other}
Below we list other types of biases that have been investigated. Without any pretension for exhaustivity, we define those biases and refer the reader to the corresponding literature for further details.
For instance, a change of functional relations can create a new bias and thus lead to $P_{train}(Y\lvert X) \ne P_{test}(Y\lvert X)$ \cite{jiang2008literature}. Also, it has been shown that biases can be created by transforming the feature space \cite{ben2007analysis}.
Furthermore, a bias specific to neural networks has been reported: During training, a change of the weights in one layer may alter the distribution of the input to the following layer. This so-called internal covariate shift slows down convergence of training a neural network and may result in a neural network that overfits \cite{ioffe2015batch}. Internal covariate shift can be compensated by normalizing the input of every layer. By doing so, it has been reported that the training can be significantly accelerated. The resulting neural network is also less likely to overfit. This approach is radically different to regularization \cite{bishop1995neural}, as it addresses the cause of overfitting rather than trying to improve a model that overfits.

\section{Conclusions}
\label{section:conclusions}
In this work, we first have presented a number of historic and modern examples of biased data sets that resulted in unreliable models. Biases occur in machine learning whenever training sets are not representative for the test data.
Even though biases have been recognized as an issue in statistics since the mid-20th century, they only recently started to get more attention in machine learning. We then provided an extensive review of biases in machine learning, with a (special) focus on the most common ones: class imbalance and covariate shift. 
We have shown how these biases can be quantified and corrected. As a consequence, in many cases it may not be helpful to simply have more data, but rather to have (possibly less) data that is more representative.

\section*{Acknowledgement}
The present project is supported by the National Research Fund, Luxembourg under grant agreement number 11508593.

\begin{footnotesize}




\bibliographystyle{unsrt}
\bibliography{References}

\end{footnotesize}


\end{document}